\RequirePackage{fix-cm}
\documentclass[smallextended]{svjour3}       
\smartqed  
\usepackage{balance}       
\usepackage{float} 
\usepackage{graphicx}
\usepackage{amsfonts}
\usepackage{xcolor}
\usepackage[hidelinks]{hyperref}
\usepackage{multirow}
\usepackage{enumitem} 
\usepackage[ruled,vlined]{algorithm2e}
\usepackage{subcaption}

\usepackage{amsmath}
\usepackage{bm} 
\usepackage{stackengine}

\usepackage{enumitem}
\usepackage{graphicx}
\usepackage{float}
\usepackage{subcaption}
\usepackage{booktabs}
\usepackage{latexsym}

\begin{document}

\title{Introducing explainable supervised machine learning into interactive feedback loops for statistical production systems}

\titlerunning{Explainable supervised machine learning at European Central Bank}        

\author{
        Carlos Mougan$^{[2]}$\and
        George Kanellos$^{[1]}$\and
        Johannes Micheler$^{[1]}$\and
        Jose Martinez Heras$^{[3]}$\and
        Thomas Gottron$^{[1]}$
}

\authorrunning{DG Statistics - European Central Bank} 

\institute{
            $^{[1]}$ Directorate of Statistics, European Central Bank\\
            \email{George.Kanellos@ecb.europa.eu, Johannes.Micheler@ecb.europa.eu, Thomas.Gottron@ecb.europa.eu}\\ 
            This paper should not be reported as representing the views of the European Central Bank (ECB). The views expressed are those of the authors and do not necessarily reflect those of the ECB.\\
            \and \\
            $^{[2]}$ University of Southampton, United Kingdom\\
            \email{C.Mougan-Navarro@soton.ac.uk}\\ 
            \and \\
            $^{[3]}$Solenix Deutschland GmbH \\
             \email{Jose.Martinez@solenix.ch}\\       
}

\date{
}

\maketitle

\begin{abstract}

Statistical production systems cover multiple steps from the collection, aggregation, and integration of data to tasks like data quality assurance and dissemination. While the context of data quality assurance is one of the most promising fields for applying machine learning, the lack of curated and labeled training data is often a limiting factor. 

The statistical production system for the Centralised Securities Database features an interactive feedback loop between data collected by the European Central Bank and data quality assurance performed by data quality managers at National Central Banks. The quality assurance feedback loop is based on a set of rule-based checks for raising exceptions, upon which the user either confirms the data or corrects an actual error. 

In this paper we use the information received from this feedback loop to optimize the exceptions presented to the National Central Banks thereby improving the quality of exceptions generated and the time spent by the users for assessing those exceptions. For this approach we make use of explainable supervised machine learning to (a) identify the types of exceptions and (b) to prioritize which exceptions are more likely to require an intervention or correction by the NCBs.
Furthermore, we provide an explainable AI taxonomy aiming to identify the different explainable AI needs that arose during the project.

\end{abstract}
\newpage

\section{Introduction}

Providing statistical data and information of the highest quality is a core task of the European System of Central Banking (ESCB). Assuring the quality of data in statistical production systems is crucial as data is used in the policy decision-making processes. Statistical production systems rely on domain experts with an outstanding understanding of the data. These experts ensure that the information used in the compilation of statistical products is of the highest quality based on their expertise and domain knowledge.

The Centralised Securities Database (CSDB)~\cite{csdb_asier} is a securities database with the aim of holding complete, accurate, consistent, and up-to-date information on all individual securities relevant for the statistical and, increasingly, non-statistical purposes of the ESCB.
Ensuring quality in such a system is challenging given the amount of data that needs to be monitored. So far, quality assurance has been achieved through a combination of data aggregation and static quality rules. In addition, \emph{business experts} with the role of Data Quality Manager use their experience for assessing data quality and manual intervention in case of issues. 

The benefit of using expert knowledge for quality assurance of statistical products is two-fold: $(i)$ all the knowledge and expertise of the domain experts is reflected in the final data and $(ii)$ the overall confidence about the final product increases on the side of the producers and consumers of official statistical data. However, as the volume and granularity of the data increase, this process becomes progressively difficult to maintain. In the case of the CSDB, which contains information on approximately 7 million alive securities, it is almost impossible to manually assess the quality of granular data without substantially increasing the number of experts.

Over the years various approaches have been developed to support and automate quality assurance procedures. However, these approaches, based on the above mentioned aggregations and static rules, have certain limitations:
\begin{itemize}
    \item \textbf{Data aggregation:} Assessing the quality of the data on an aggregated level condenses the information that needs to be checked to a manageable amount. The disadvantage of this approach is that aggregation might hide granular data quality issues on the level of individual observations. The risk of obfuscation depends on the type and level of aggregation used.
    \item \textbf{Static quality rules:} Another way to ensure data quality for big data, is using a fixed set of rules to identify potential quality issues. These rules are based on expert knowledge and formalize certain aspects of their domain knowledge. Such rules flag individual records that, for instance, exceed a predefined threshold. The flagged observations typically represent only a small fraction of the entire data set, allowing the field experts to focus their analysis on a limited amount of data. The shortcoming of this approach is that relying on a fixed set of static rules is not flexible enough to capture dynamics in data nor does it automatically adjust to new insights provided by users. 
\end{itemize}

In this paper we make the following contributions:
\begin{enumerate}[label=(\roman*)]
\item  First, we identify and describe the types of actions that experts have been performing historically in the context of their data quality assurance tasks. This interaction forms the basis for our feedback loop. The information obtained in this way permits the formulation of detecting data quality issues  as a supervised machine learning problem. The concrete objective is to predict the probability that a granular data instance is an outlier.

\item  Once we are able to identify the probability of a data quality issue, we process the results to produce a set of ranked exceptions. The ranking is based on the product of the amount outstanding or market capitalization for a certain financial instrument multiplied by the soft prediction of the machine learning model that the observation is an outlier. This ranking allows data quality managers to focus on those observations with a high risk of being wrong while keeping business priorities in mind.

\item Furthermore, modern practices for machine learning models require $(i)$ high predictive accuracy and $(ii)$ model interpretability \cite{desiderataECB}. One of the barriers that artificial intelligence (AI) is facing regarding practical implementation in the financial and public policy sectors is the inability to explain or to fully understand the reasons why an algorithm takes certain decisions \cite{xai_concepts,rudin2019stop}. With both reasons in mind, in Section \ref{explainability}, we provide desiderata for explainability in AI by categorising users and analysing potential needs.

\end{enumerate}

The rest of the paper is organized as follows: First, we give an overview of the data and existing data quality assurance processes and tools. In Section~\ref{sec:solution} we describe our approach, including the data pre-processing and feature engineering steps, the methods used to solve the given task and an evaluation of performance. Finally, we provide a set of explanability needs for the model before concluding the paper with a summary and a review of current limitations towards future work.
\section{Dataset Overview}
The Centralised Securities Database (CSDB) \cite{csdb_asier} is a securities database with the aim of holding complete, accurate, consistent, and up-to-date information on all individual securities relevant for the statistical and, increasingly, non-statistical purposes of the European System of Central Banks (ESCB).

It is a single information technology infrastructure that contains master data on securities (e.g. outstanding amounts, issue and maturity dates, type of security, coupon and dividend information, statistical classifications, etc.), issuers (identifiers, name, country of residence, economic sector, etc.) and prices (market, estimated or defaulted) as well as information on ratings (covering securities, issuance programs, and all rated institutions independently of whether they are issuers of securities). 

The CSDB covers securities issued by EU residents; securities likely to be held and transacted in by EU residents; and securities denominated in Euro, regardless of the residency of the issuer and holders. The CSDB currently contains information on over 7 million non-matured or “alive” debt securities, equities, and mutual fund shares/units plus approximately forty-nine million matured or “non-alive” (e.g. matured, early redeemed, or canceled) securities. 

Developed by the ECB, the CSDB is jointly operated by the members of the ESCB and it is only accessible by them, i.e. it is not available for public purposes. The CSDB uses data from commercial data providers, national central banks, and other sources. The most plausible value for each attribute and instrument is selected through a weight-based algorithm and gaps of missing information (in particular for prices and income) are filled with reliable estimates. The system makes use of expertise within the ESCB to enhance data quality.

The CSDB provides consistent results and harmonization of concepts and calculations for all users together with efficiency in the data reporting, which reduces the burden from reporting agents, and improves the data compilation process.

\subsection{iDQM Tool}
Data quality management in the CSDB is based on the DQM framework \footnote{\url{https://eur-lex.europa.eu/eli/guideline/2012/689/oj}} that was established in 2012 and provided the guidelines for assessing data quality. Monitoring was performed once a month by using data snapshots. These snapshots focused on different dimensions such as changes in individual attributes, problems in issuer identification, and comparison with benchmark statistics, where certain target thresholds had to be fulfilled. 
Monitoring was performed via a Business Intelligence tool which allowed for data exploration but did not offer the possibility to amend or confirm potential issues. The only possibility to address them was via other tools and these changes were not reflected in the data snapshots.

To tackle these issues, a new tool was developed which allowed an interactive approach to data quality management: the iDQM. 
This tool visualises exceptions in the data that might require manual verification and potential intervention. Furthermore, the iDQM tool links the transactional part of the database (where all the data processing and calculations take place) with the data warehouse (where all the snapshots are taken from) and offers the possibility for data quality managers to amend errors and update the target metrics instantaneously.   
Moreover, the iDQM offers better prioritization in the exception generation, including an audit tool for tracing actions and confirmations.

\subsection{Bulk Tool}

The complementary bulk tool enables Data Quality Managers to change one or more instrument attributes directly. This tool is used for interventions on a larger scale and can be used to correct multiple values at the same time. These corrections do not necessarily correspond to detected exceptions. 

\subsection{Audit Records}

Any data change performed either through the iDQM or the Bulk tool creates an audit record.
Audit records contain information on how the data was before and after the change. By extracting a historical log of audit records we obtained a dataset of over 1 million changes for 25 different exception types.
Please note, that the majority of the records were based on corrections made via the bulk tool.






\section{Our Solution}\label{sec:solution}

Our objective is to identify data records in CSDB which might require a correction by a Data Quality Manager. 
Rather than explicitly modelling domain knowledge as static rules for identifying such records, we want to leverage the experts' knowledge implicitly contained in past data corrections. This means, we want use historic interventions of Data Quality Managers to generalise the patterns underlying the corrections of data records. These patterns can then be used to predict whether new data records might require a manual quality check, to prioritise the detected issues and to ideally propose which value requires an adjustment.

These objectives can be translated into three distinct tasks:
\begin{enumerate}[label=(\roman*)]
    \item Identify if a record might need to be checked and potentially changed by a Data Quality Manager.

    \item Rank these suggestions to increase efficiency of the Data Quality Manager in reviewing data records by considering both: the economic relevance of an instrument and the probability that an observation is an outlier.
    
    \item Propose to the Data Quality Manager which is the field that is most likely to be incorrect.
\end{enumerate}


\subsection{Feature Engineering}

The initial and very important step before modeling a machine learning prediction problem is to convert raw data into meaningful features. 
The features must capture information about the data that might be of relevance for the learning process. Only if the features actually capture the signals indicating data quality issues, a machine learning approach will be capable to learn from the data and achieve a good generalisation and predictive performance.
This is why the steps of data preparation and feature engineering are crucial in the life cycle of any machine learning project ~\cite{mle,quantile2021}. In our particular case we created the following features:

\begin{itemize}
    \item \textbf{Percentage change for numerical columns}:  The relative change compared to the month before the prediction. Additional to this feature, we created the relative change to the median of the last three months, which is sometimes more reliable due to the statistical robustness of the median.
    
    \item \textbf{Lag features}: Boolean features denoting if the value of a column has been changed with respect to the previous month. 
    
    \item \textbf{Time difference for date columns}:  Date columns can lead to overfitting since the model might learn an exact date rather than a pattern. Instead of using dates or timestamps, they are replaced by the time difference between two events.
    
    \item \textbf{Categorical columns}, A high ratio of the problem features are categorical features with a high cardinality. To utilize these features in our model we used Target Encoders \cite{quantile2021,Pargent:2019,high-cardinality-categorical} with regularization to avoid overfitting or data leakage \cite{mougan2022fairness,pargent2021regularized},
\end{itemize}

\subsection{Evaluation Setup}

To evaluate and assess how well a model generalises and predict how it will perform in a production environment it is necessary to split the available data in subsets used for training, validating and testing the model.
Given the setup of the CSDB system and the intended use of the classification model we decided to split the data along the temporal dimension. This means we used the historically oldest data for training the model, and more recent data for validation and testing. A temporal split corresponds to the envisaged use case of applying the model to novel observations, arriving in the system after the model has been trained and deployed.
This also addresses the specific challenge of distribution variability across different types of exceptions over the time axis.
Our split is made following a ratio of 60:20:20 for train, validation and test data.

Furthermore we created an additional gold-standard subset of our test data. This gold-standard subset was constrained to those exceptions that were corrected via the iDQM tool. The motivation for this additional evaluation dataset was to see how well the models trained on the overall corpus would perform in the context of the iDQM tool, where the data quality managers typically perform their work.

For the sub-tasks mentioned above, we used different evaluation metrics:

\begin{enumerate}[label=(\roman*)]
    \item The learning task to detect which data records require a manual intervention corresponds to a binary classification problem. Hence, we use classical metrics based on a confusion matrix. By denoting true positive results as $TP$, false positive as $FP$ and false negative results as $FN$ we make use of the precision and recall metrics.
    \begin{itemize}
        \item \textbf{Precision} is a classification metric that measures the quality of the prediction. It is defined as:
        
        \begin{equation}
        \textit{precision} = \frac{TP}{TP + FP}
        \end{equation}
        
        In our case, precision tells us for which percentage of the cases the machine learning model was right when it predicted that a Data Quality Manager would change a record. 

        \item \textbf{Recall} is a classification metric that measures the percentage of positive instances that were identified by the model, defined as:
        
        \begin{equation}
        \textit{recall} = \frac{TP}{TP + FN}
        \end{equation}
        
        In our case, recall tells us for which percentage of all the cases where a Data Quality Manager has actually changed a value, was the machine learning model able to predict this need of a change. 
    \end{itemize}
    
    \item The task of ordering the identified data records in such a way that a data quality manager encounters mainly items in the top positions which require an intervention is a ranking task.
    
    As evaluation metric we employ the normalized discounted cumulative gain (NDCG). NDCG is, as its name suggests, a normalized version of the discounted cumulative gain (DCG). The traditional formula of DCG accumulated at a particular rank position $p$ is defined as:
    \begin{equation}
     DCG_p =\sum_{i=1}^{p}\frac{rel_i}{\log_{2} (i+1)}\\
    \end{equation}
    \begin{equation}
    NDCG_{p}=\frac{DCG_p}{IDCG_p}
    \end{equation}\label{eq:ndcg}
    where $IDCG_p$  is the ideal discounted cumulative gain, i.e. the cumulative gain value for an optimal ordering. Typically NDCG is computed for certain cut-off values K in the ranking, assuming a human user will not look at all items, but only at the top-K positions.

\end{enumerate}

\subsection{Predicting outliers}

In the context of this project we use supervised classification methods to predict the probability that a data point will need to be amended by a Data Quality Manager. This corresponds to detecting outliers and can be modelled effectively as a binary classification task.

The methods we tried out for this problem were logistic regression, decision trees, support vector machines and gradient boosting (through CatBoost). For all methods we employed the standard Python implementations available in the \texttt{scikit-learn} package ~\cite{scikit-learn}. 

We found that not all of the 25 different exception types we observed in our data set were frequent enough to train and evaluate the machine learning methods. The other exceptions occurred too infrequently to provide sufficient data for inferring a generalised classification model. Furthermore, very early in the process it turned out that the CatBoost algorithm significantly outperformed the other approaches.

In Table~\ref{tab:detection} we present the performance of CatBoost on the full test dataset in terms of precision and recall for detecting different types of exceptions. What can be seen is that the performance varies across different types of exceptions. This is an indication that the patterns underlying those exceptions might be quite different and accordingly more or less challenging to detect.

\begin{table}[ht]
\centering
\begin{tabular}{l|llll}
\textbf{Exception Type}         & \textbf{Precision} & \textbf{Recall} \\ 
\cline{1-3}
\textit{AmountOutstanding}       & 0.409         & 0.762            \\
\textit{CouponDate}         & 0.063           & 0.305             \\
\textit{SecurityStatus}         & 0.456         & 0.801            \\
\textit{MaturityDate}           & 0.409           & 0.505              \\
\textit{IssueDate}          & 0.963              & 0.667            \\
\textit{DividendAmount}          & 0.369              & 0.551            \\
\textit{ESAI2010}          & 0.735              & 0.996            \\
\end{tabular}
\caption{Model performance of CatBoost across different types of anomaly detection on the overall test dataset.}
\label{tab:detection}
\end{table}

To assess how suitable the models are for deploying them in the iDQM tool we additionally checked their performance on the gold-standard data set. 
The first observation we made was that only seven of the exception types appeared in the gold-standard data set. This might be explained by Data Quality Managers addressing only certain types of exceptions in the iDQM tool. For other types of exceptions, which might indicate some systematic errors, they use the bulk tool for correcting larger amounts of data records.

Furthermore, as shown in Table~\ref{tab:detection-gold} the performance of the models differs quite a lot from the full data set. For some exceptions type the precision and recall for the binary classification task drop to 0, which means that no data record was marked as being an outlier.
This deviation in the performance indicates that the two data sets behave quite differently and 
that the observations from the iDQM tool follow a different distribution compared to the values from the more frequently used bulk tool.
However, there still is some signal from the training data which can be leveraged to identify data records which might require a manual check and provide a relevance based ranking. 

\begin{table}[ht]
\centering
\begin{tabular}{l|llll}
\textbf{Exception Type Precision Recall}  & \textbf{Precision} & \textbf{Recall} &  &  \\ \cline{1-3}
\textit{AmountOutstanding}       & 0.502              & 0.230            &  &  \\
\textit{CouponDate}             & 0.000              & 0.000               &  &  \\
\textit{SecurityStatus}         & 0.455              & 0.746           &  & 
\\
\textit{MaturityDate}           & 0.000                 & 0.000              &  & \\
\textit{IssueDate}          & 0.000              & 0.000           &  &  \\
\textit{DividendAmount}          & 0.500              & 0.621           &  &  \\
\textit{ESAI2010}          & 0.249              & 0.365           &  &  \\
\end{tabular}
\caption{Model performance of CatBoost  on the iDQM specific gold-standard dataset}
\label{tab:detection-gold}
\end{table}

\subsection{Ranking Problem}
Ultimately, the objective is to produce a ranking in order to prioritize the records that Data Quality Managers would need to investigate. Ideally, the order in this ranking would maximize the number of true positives in higher positions and reduce the amount of work for the Data Quality Managers.
 
 The ranking for a certain record $(R_p)$ is based on the product of the amount outstanding or market capitalization $(\texttt{AO}_p)$ for a certain financial instrument multiplied by the soft prediction of the machine learning model that the observation is an outlier $(f(x_p))$ (cf. Equation~\ref{eq:rank}).
In this way we combine the insights from the machine learning model with the business priority to ensure correctness of data entries which are of higher economic relevance.

\begin{equation}
    R_p = f(x_p) \cdot \texttt{AO}_p \label{eq:rank}
\end{equation}

Table~\ref{tab:ndcg-gold} shows the NDCG values for ranking the identified exceptions following equation (\ref{eq:rank}). Ranking order is crucial for user experience in the context of the iDQM as it instills trust in the ranking algorithm and ensures that the limited resources dedicated to data quality will be focused on the most pertinent data issues. 
Applying the ranking brings several of the identified cases close to the top of the results, causing the data quality managers to be primarily exposed to those cases which actually require some manual intervention.

\begin{table}[ht]
\centering
\begin{tabular}{l|llll}
\textbf{Exception Type}                   & \textbf{K=10} & \textbf{K=50} & \textbf{K=100} & \textbf{K=1000} \\ \hline
\textit{AmountOutstanding}      & 0.773         & 0.781         & 0.782          & 0.770           \\
\textit{CouponDate}             & 0.969         & 0.994         & 0.994          & 0.994           \\
\textit{SecurityStatus}         & 0.930         & 0.913         & 0.911          & 0.963           \\
\textit{MaturityDate}           & 0.500         & 0.500         & 0.543          & 0.843           \\
\textit{IssueDate}              & 0.395         & 0.664         & 0.664          & 0.664           \\
\textit{DividendAmount}              & 0.000         & 0.000         & 0.000          & 0.000           \\
\textit{ESAI2010}              & 0.000         & 0.000         & 0.000          & 0.000           \\
     \end{tabular}
\caption{NDCG ranking results on gold dataset for different exception types and ranking cut off values K.}
\label{tab:ndcg-gold}
\end{table}

\section{Explanations Taxonomy}\label{explainability}
In this section we first overview the different types of users that interact with the CSDB and expand on their profiles. Afterwards, we analyse the different explainable AI desiderata that arise through ML implementation in a statistical production system. For more details about the desiderata of explainable AI in statistical production systems we suggest~\cite{desiderataECB}.

\subsection{Users}

During the project, we came across several generic user roles which help to classify the needs for solutions of explainable and responsible AI. A key question driving this classification is \emph{Who needs an explanation of an AI method?} This helps to clearly define and distinguish different desiderata for explanations. 

The following profiles define users that can potentially interact with a machine learning system all of whom have quite different needs for explainability:

\begin{enumerate}[label=(\roman*)]
    \item \textbf{Data scientists and AI engineers}: This role corresponds to members of the team who build, model, and run a machine learning application. This type of user has technical expertise but does not necessarily have business expertise. They are in charge of the full life cycle of the machine learning application, from development to maintenance in production.

    \item \textbf{Business experts}: Users with this role provide the use case and domain expertise for a machine learning solution. They define the business activity or process which is supported by the AI solution. In our case, they are finance, economics, and statistics experts from the European System of Central Banks who act or intervene in business processes based on the recommendations of the models. This type of user might not have a technical background and is not a machine learning expert.
    
    \item \textbf{High stake decision makers}: This type of user determines whether to use and incorporate a machine learning model in the decision-making process. They typically have a management position, a high-level understanding of the business objectives, and a responsibility to deliver value. They need to understand and assess the potential risk and impact of incorporating the machine learning model into production.
    
    \item \textbf{End users}: Users which are affected by or make use of the final results belong to the group of end-users. The knowledge and potential expertise of this user group vary significantly. There might be cases where the group of end-users overlaps or is even identical to the group of \emph{business experts}, e.g. when the machine learning solution is primarily serving internal business processes. Examples of end-users in our domain are the business areas or even the general public making use of data compiled by the Directorate General Statistics at the ECB.
\end{enumerate}

\subsection{Building Trust}

Trust is essential for the adoption of a machine learning application. Depending on the user, the meaning of trust and the way to obtain it differ\cite{desiderataECB}. Despite the careful testing and calibration of the machine learning process by the \emph{data scientists}, \emph{end users} of the data can identify potential data issues that have not been raised by the algorithm. These issues are communicated via the Data Quality Managers to the statistical production team (comprising of \emph{business experts} and \emph{data scientists}) responsible for the CSDB data quality.

\begin{figure}[tb]
\begin{subfigure}{.5\textwidth}
  \centering
  \includegraphics[width=.8\linewidth]{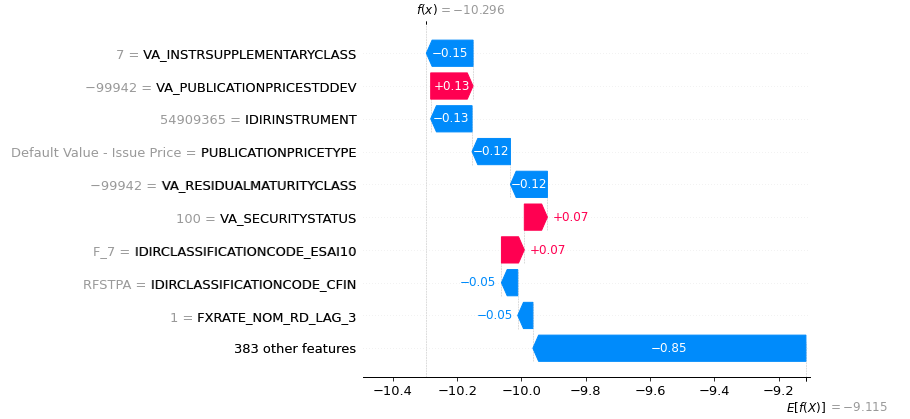}
  \caption{Local feature importance for a given instance\cite{shapley}}
  \label{fig:sub-first}
\end{subfigure}
\begin{subfigure}{.5\textwidth}
  \centering
  \includegraphics[width=.8\linewidth]{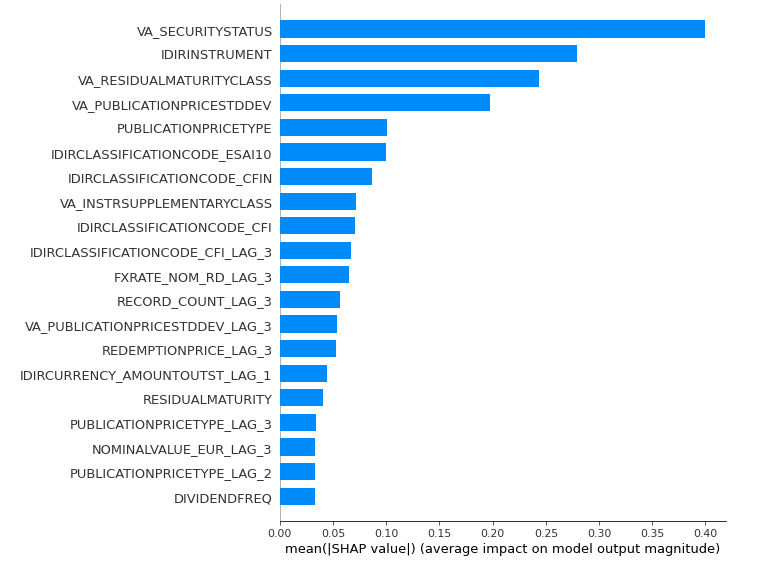}
  \caption{Global Feature Relevance}
  \label{fig:sub-second}
\end{subfigure}
\caption{Shapley feature values by TreeExplainer~\cite{shapTree,lundberg2020local2global,lundberg2017unified}. Feature relevance explanations are useful to detect data leakage and prevent undesired model behaviour.}
\label{fig:fig}
\end{figure}

The first step for the statistical production team is to verify that indeed the issues identified are not flagged by the algorithm. The next step is to investigate the reasoning behind this choice from the perspective of the algorithm by determining: $(i)$ similar instances in the training dataset~\cite{case_based_explanation}, $(ii)$ which features contributed to the decision~\cite{shapTree,ribeiro2016why} (cf. Figure~\ref{fig:fig}) and $(iii)$ logical decision rules to better understand the model logic~\cite{lore,Ribeiro2018AnchorsHM,State21}.

\subsection{Actionable Insights}

Understanding a machine learning algorithm is usually not an end in itself. The explanation offered through this understanding supports business processes and leads to actionable insights. Such insights enable the \emph{business expert} to understand how to change a decision by manually intervening in the data. 
For instance, when data is identified to belong to a certain class, providing a set of actionable changes that would lead to a different decision can assist an expert in correcting or modifying the data. 

Counterfactual generation aims to address this issue by proposing to the \emph{business experts} \emph{the minimal feasible change in the data in order to change the output of the algorithm}. Such a process enhances the understanding of the experts (and might further foster trust in the system). We formulate the problem following the counterfactual recourse formulation by Ustun et~al.~\cite{ActionableRecouserLinear,AlgRecourse_Counterfactual_Interventions} and the open source python package DICE ~\cite{dice} that quantifies the relative difficulty in changing a feature via feature weights.

We quantified the relative difficulty in changing a feature through the application of weights to the counterfactual explanations algorithm. For instance, in our case, recommendations should not ask the \emph{business expert} (Data Quality Manager) to modify the country in which a financial instrument was issued or change the issue date to a time before the creation of the issuing company (cf. Figure~\ref{fig:counterfactual}).

\begin{table}[ht]
\centering
\begin{tabular}{l|cccc}
\textbf{Feature}     & \textbf{Original Value} & \textbf{Modified Value} & \textbf{Initial Pred.} & \textbf{Modified Pred.} \\ \hline
VA SecurityStatus    & 100               & 201               & 1                           & 0                            \\
CODE ESAI10          & $F\_31$          & $F\_32$           & 0                           & 1                            \\
VA SecurityStatus    & 101               & 203               & 0                           & 1                            \\
PublicationPriceType & CLC               & PAY               & 1                           & 1                           
\end{tabular}
\caption{Set of counterfactual decisions generated~\cite{lore,dice}. Counterfactual explanations can be understood as \emph{What is the minimal feasible change in the data in order to change the output of the algorithm?}}
\label{fig:counterfactual}
\end{table}

\subsection{Model Monitoring}

Model monitoring aims to ensure that a machine learning application in a production environment displays consistent behavior over time. Monitoring is mainly performed by the \emph{data scientist and AI engineer} and is crucial, as a drop in model performance will affect all the users. Two common challenges in model monitoring are $(i)$ distribution shifts in the input data that can degrade model performance $(ii)$ changes in the machine learning algorithm due to a model retraining that can alter the individual explanations for decisions.

Detecting when the underlying distribution of the data changes is paramount for this use case, since failing to predict outliers or errors in the data will lead to a drop in the trust of the machine learning model. Also, the risk of having an incoherent explanation through time caused by the continual learning process is important as a discrepancy will lead to a decrease of trust by the \emph{business experts}.

Diverse types of model monitoring scenarios require different supervision techniques. We can distinguish two main groups: Supervised learning and unsupervised learning. Supervised learning is the appealing one from a monitoring perspective, where performance metrics can easily be tracked. Whilst attractive, these techniques are often unfeasible as they rely either on having ground truth labeled data available or maintaining a hold-out set, which leaves the challenge of how to monitor ML models to the realm of unsupervised learning~\cite{continual_learning,ShiftsData,mougan2022monitoring,garg2022leveraging}.

In this work since we are in the realm of unsupervised learning we have considered two possible solutions:
\begin{itemize}
    \item Obtaining model uncertainty estimates via non-parametric bootstrap as an indicator of model performance when the deployed data is not available~\cite{mougan2022monitoring}. This monitoring technique allows to have an indicator of the model performance, detecting increases uncertainty lead to identifying indicators of when the model performance deteriorates in unsupervised data scenarios.
    
    \item Another approach suggested by Lundberg et~al.~\cite{shapTree} is to monitor the SHAP value contribution of input features over time together with decomposing the loss function across input features in order to identify possible bugs in the pipeline.
\end{itemize}

\subsection{Fostering Explanations through Simple Models}

Copying \cite{copies,copying_irene_sampling} or distilling \cite{Hinton_distill}  machine learning models can greatly contribute to model explainability. Overly complex models tend to be difficult to explain~\cite{mle} and can become unaccountable~\cite{rudin2019stop}. Model agnostic copies with a simple model might be able to achieve global explainability~\cite{unceta2018global} which can be useful to build trust and gain knowledge by the \emph{business expert}. Furthermore, in some deployment scenarios involving incompatible research and deployment versions~\cite{copies}, copying the ML model can ease the deployment task for the \emph{data scientist}. 

For our case, the original model $f_O(X)$ is a CatBoost classifier~\cite{catboost-encoder} which is a gradient boosting model ~\cite{gbm} that often achieves state of the art results in many different types of problems and the copied models $f_C(X)$ are a scikit-learn~\cite{scikit-learn} decision tree classifier and a Generalized Linear Model (GLM). The simpler models are of slightly inferior quality  (cf. Table~\ref{tab:copy}). However, having two simple models helps to improve the overall global explainability for the \emph{data scientist} and to simplify deployment.

\begin{table}[bt]
\centering

\begin{tabular}{lccc}
\toprule
          & Catboost & Decision Tree & GLM \\ 
\midrule
AUC       & 0.816    & 0.771         & 0.741        \\
Precision & 0.805    & 0.741         & 0.685        \\
Recall    & 0.833    & 0.824         & 0.733   
\\
\bottomrule
\end{tabular}
\caption{CatBoost is the original model $f_O(X)$ and the Decision Tree and the Generalized Linear Model the copied classifier $f_C(X)$. The creation of model surrogates or copies can distil the knowledge in the machine learning process thereby improving explainability.}\label{tab:copy} 
\end{table}

\section{Conclusions}

Statistical production systems are one of the most promising fields for the adoption of machine learning processes within the context of central banking. Its maturity is still in the very early stages. In this paper, we have introduced our approach to improving the data quality exceptions presented to the Data Quality Managers of the Centralised Securities Database. This approach aims to first identify potential exceptions and then rank them based on the probability that they would need manual intervention by a Data Quality Manager.

Due to the importance of the quality assurance process and following the requirements of modern responsible machine learning, and explainable machine learning pipeline is needed to support the stakeholders involved in the data quality assurance task. With this reason in mind, a series of potential Machine Learning accountability desiderata have been presented tackling possible explainability needs that may arise.

The correct analysis, development, and introduction of machine learning in statistical production systems can lead to an optimization of the interventions needed to maintain data quality thereby making efficient usage of the limited resources available. This efficiency gain can translate into reduced time effort for Data Quality Managers or an increase in data quality. Eventually and over time such an ML pipeline can also provide new insights into the data as well as greater trust in the underlying processes.

\section*{Limitations}

Using Machine Learning solutions for data quality assurance processes in granular statistical data collections is a very promising approach. An important insight from our project on the CSDB illustrates the potential of such approaches but also highlights some of the limitations.

Most importantly, the availability of sufficient and high-quality training data is crucial for successfully employing supervised machine learning methods. 
The volume, characteristics, and information contained in a data set with labeled outliers are essential for the quality of a trained model. We have seen in our case, that if the data used for training deviates from the data observed in a production setting this can negatively influence the performance of the trained model. This also underlines the importance of a sound evaluation methodology. The setup we chose allowed us to detect some limitations of the performance early on and before moving a full-fledged system to a production environment.

In conclusion, the main limiting factor currently is too short historicity of interventions of data quality managers for most of the exception types. While for some cases of exceptions the data permitted to train models of sufficient quality, many exceptions suffered from sparsity in the training models. 

Furthermore, the errors corrected using the bulk tool deviate to a certain extent from the corrections made in iDQM. These similar but distinct feedback loops might serve as the basis for two different types of machine learning tasks, addressing different types of quality issues. Transferring the insights from one tool to the other seems to provide only limited insights.

\section*{Acknowledgements}
This work was partially funded by the  European  Commission under contract numbers NoBIAS — H2020-MSCA-ITN-2019 project GA No. 860630. \\



\bibliographystyle{spmpsci}      
\bibliography{bib}
\end{document}